\documentclass[11pt,a4paper]{article}
\usepackage[hyperref]{acl2020}
\usepackage{times}
\usepackage{latexsym}
\usepackage{amssymb}
\usepackage{amsmath}
\usepackage{commath}
\usepackage{bm}

\usepackage{graphicx}
\usepackage{booktabs}
\usepackage{tabularx}
\newcolumntype{R}{>{\centering\arraybackslash}X}
\newcommand*{\leftHead}[1]{\multicolumn{1}{>{\raggedright\arraybackslash}X}{#1}}

\usepackage{pifont}
\usepackage{microtype}

\aclfinalcopy 

\title{A Survey on Semantic Parsing from the Perspective of Compositionality}
\author{Pawan Kumar \\
  Indian Institute of Technology Delhi, \\
  New Delhi, Hauzkhas-110016 \\
  \texttt{pawan.kumar@cse.iitd.ac.in}\\\And
 Srikanta Bedathur \\
 Indian Institute of Technology Delhi, \\
 New Delhi, Hauzkhas-110016 \\
 \texttt{srikanta@cse.iitd.ac.in} \\
}

\date{17th July 2020}

\begin{document}
\maketitle
\begin{abstract}

Different from previous surveys in semantic parsing \citep{kamath2018survey} and knowledge base question answering(KBQA) \citep{chakraborty2019introduction, zhu2019statistical, hoffner2017survey} we try to takes a different perspective on the study of semantic parsing. Specifically, we will focus on (a) meaning composition from syntactical structure ~\citep{partee1975MontagueGrammar}, and (b) the ability of semantic parsers to handle lexical variation given the context of a knowledge base (KB). 
In the following section after an introduction of the field of semantic parsing and its uses in KBQA, we will describe meaning representation using grammar formalism CCG \citep{steedman1996surface}. We will discuss semantic composition using formal languages in \ref{sec:compositionality_in_formal_semantics}. In section \ref{sec:logical-form} we will consider systems that uses formal languages e.g. $\lambda$-calculus \cite{steedman1996surface}, $\lambda$-DCS \cite{liang2013lambda-DCS}. Section \ref{sec:graph-form} and \ref{sec:tree-form} consider semantic parser using structured-language for logical form.  Section \ref{sec:benchmark} is on different benchmark dataset ComplexQuestions \citep{bao2016ComplexQuestions} and GraphQuestions \citep{su2016GraphQuestion} that can be used to evaluate semantic parser on their ability to answer complex questions that are highly compositional in nature.
\end{abstract}

\section{Introduction}
One of the main challenge in Knowledge Base Question Answering (KBQA) is semantic parsing - the construction of a complete, formal, symbolic, meaning representation (MR) of a sentence \citep{wong2006learning}. Most commonly used formal frameworks use a combination of $\lambda$-calculus and first order logic (FOL) e.g. CCG \citep{zettlemoyer2005ccg}, $\lambda$-DCS \citep{liang2013lambda-DCS}. The logical-expression further needs to be grounded in a knowledge base (KB),  in the case of KBQA, the challenge here is lexical variation. The two main challenges that any KBQA system has to tackle are language compositionality: evident by the choice of the formal language and the construction mechanism and lexical variation: grounding the words/phrases to appropriate KB entity/relation.

\paragraph{Language Compositionality:} According to Frege's principle of compositionality: ``the meaning of a whole is a function of the meaning of the parts. The study of formal semantics for natural language to appropriately represent sentence meaning has a long history \citep{sep-compositionality}. Most semantic formalism follow the tradition of Montague's grammar \citep{partee1975MontagueGrammar} i.e. there is a one-to-one correspondence between syntax and semantics e.g. CCG \citep{zettlemoyer2005ccg}. We will not be delving into the concept of semantic representation in formal language in this survey.   

\paragraph{Lexical Variance:} Lexical variation in human language is huge. Differences in the surface form of words in the natural language and the label of the corresponding entity/relation in the KB is mainly due to the polysemy. For example \textit{attend} may be referred to by label \textit{Education} \citep{berant2013semantic}. Similarly paraphrases of a sentence may have different phrases to mean the same thing, e.g. `What is your \textit{profession}', `What do you \textit{do for a living}'. `What is your \textit{source of earning}' all these variation may points to label \textit{profession} \citep{berant2014pp}. 

The two challenges are elegantly summed up in a function $p = f(a, b, R, K)$ by \citet{mitchell2010composition} i.e. the combined meaning of symbol a and b is function of lexicon a and lexicon b under the syntactic relation $R$ and the context $K$. We propose that the context $K$ be the knowledge base (KB). KBQA provides an appropriate ground for testing different semantic parsing approaches empirically. There are some KBQA systems which use semantic parser as a module in their pipeline \citep{reddy-etal-2014-large, cheng-reddy-2017-tree_transformation_rule} where the purview of semantic parsing is to get to the logical-expression, and a downstream process takes up lexicon grounding or disambiguation. There are some systems which don't have such seperations e.g. SEMPRE\citep{berant2013semantic}. However, both type of systems do resort to some formal language or intermediate logical form. We exclude KBQA systems which use non-symbolic representation \citep{cohen2020reifKB}.

We describe here some terminology commonly used in the study of semantic parsing \citep{diefenbach2018core, kamath2018survey}. 1. Intermediate logical form: represents the complete meaning of the natural language using formal language e.g. $\lambda$-calculus \citep{zettlemoyer2005ccg, hakimov2015lexgap}, $\lambda$-DCS \citep{liang2013lambda-DCS} or structured language \citet{Scott_yih_and_Jianfeng_2015_SQG-Tx-KB,sen_hu_emnlp_2018SQG-Tx-KB}. This is the main output of semantic parsing. 
2. phrase-mapping: mapping phrases in the question to their corresponding resource in the KB is required to provide a real-word context to the intermediate logical form. This process is also called grounding of the logical form, thus obtaining a \textit{grounded logical form}. 
3. Disambiguation: of many resources obtained in the phrase-mapping process only a few will be right according to the semantic of the natural language.
4. Query construction: Querying the KB-endpoint requires translation of the grounded logical form into a query language e.g. SPARQL. Translation from grounded logical form to the query language is a deterministic process.

\section{Compositionality in Formal Semantics}
\label{sec:compositionality_in_formal_semantics}
Considering the compositionality in the natural language in the sense that meaning of the whole sentence is constructed from meaning of its parts. According to \citet{Jeff_Pelletier} this is a compositionality in the ``functional sense": something is compositional if it is a complex thing with some property that can be defined in terms of a function of the same property of its parts (with due consideration to the way the parts are combined). In formal semantics the complex things is a syntactically complex sentence and the property of interest is meaning, while combining the parts due consideration has to be given to how those parts are syntactically present in the complex sentence. The choice of a formal language to represent meaning of a complex sentence in a system trying to parse the natural language sentence into functional composition of meaningful parts greatly affects its capacity (expressiveness). Such a system is known as Semantic Parser.

\subsection{Formal Language} First order predicate logic (FOPL) can be used to represent meaning of natural language sentence, however it fails to represent some concepts in the natural language, e.g. ``How many primes are less than 10?" \citep{liang_fol_v_lambda} - FOPL doesn't have a function to count the number of elements. The formal semantics can use a higher-order language e.g. $\lambda$-calculus, say a higher order function $count$ exists, that can count the number of elements in a set. Thus we can represent the previous questions as $count(\lambda x.prime(x)\wedge less(x, 10))$. Without going into further details of the formal language, let's consider an example here showing compositional use of $\lambda$-calculus to represent the meaning of a complex sentence. E.g. the sentence ``Those who had children born in Seattle." \citet{liang2013lambda-DCS}.
\begin{align*}
\lambda x.\exists y.children(x, y) \wedge PlaceOfBirth(y, Seattle)
\end{align*}
Take another example showing coordination in $\lambda$-calculus "Sqaure blue or round yellow pillow" \citep{artzi2013LC_using_CCG} which is represented as
\begin{align*}
\lambda x.pillow(x)\wedge \big((square(x)\wedge blue(x)) \vee \\ (round(x)\wedge yellow(x))\big)
\end{align*}
Many semantic parsing systems use only part of the operators available in $\lambda$-calculus thus they are limited in expressiveness by their choice of operators, not by the choice of formal language. 
  
\subsection{Structured language} 
\begin{figure}[h]
    \centering
    \includegraphics[scale=0.5]{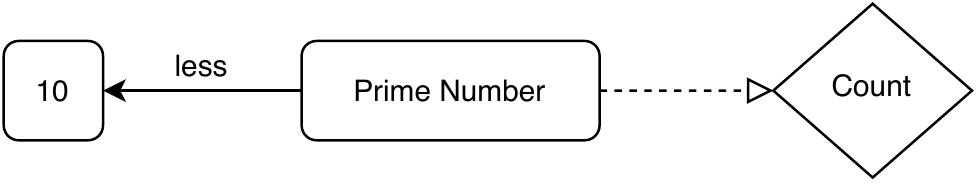}
    \caption{logical form for ``How many primes are less than 10?" as graph-structure}
    \label{fig:sqg_example}
\end{figure}

A graph-structured logical form or a tree-structured logical form can also be used to represent the meaning of a natural language sentence. 
\paragraph{Graph-structured language} E.g. Semantic Query Graph(SQG) \cite{sen_hu_ieee-2018_NFF}, has nodes representing constant/values and edges representing relation. The edges could be seen as analogous to the binary relation of the logical formalism. This definition of SQG directly corresponds to the many knowledge graphs $\mathcal{K}$ like dbpedia \citep{auer2007dbpedia}, freebase \citep{bollacker2008freebase}. With four primitive operations to manipulate a graph structure: connect and merge that operate on pair of nodes,  expand and fold that operate on single node \citep{sen_hu_emnlp_2018SQG-Tx-KB}, and with higher-order functions attached to nodes \citep{Scott_yih_and_Jianfeng_2015_SQG-Tx-KB}, graph-structure makes for good candidate for logical form of a natural language sentence e.g. Figure \ref{fig:sqg_example}.

\begin{figure}
    \centering
    \includegraphics[scale=0.4]{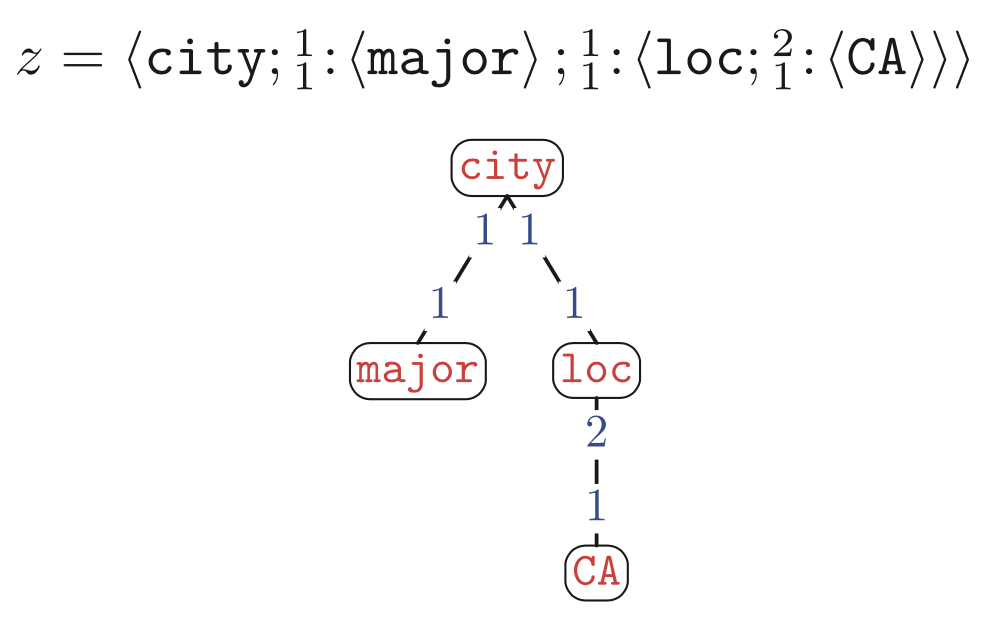}
    \caption{DCS tree for sentence ``Major city in California"}
    \label{fig:dcs_tree}
\end{figure}

\paragraph{Tree-structured language} Logical languages with tree-hierarchy can represent such hierarchy of natural language as well. FunQL \citet{cheng-reddy-2017-tree_transformation_rule, zelle1996chill_ilp, kate2005Slit_transformation_to_tree} is a variable free functional language encoding tree-hierarchy. It has a predicate-arguments form and a recursive tree structure, where the non-terminal are predicate and the terminal nodes in the tree makes for an argument, e.g. sentence ``which states do not border texas?"  is:
\begin{align*}
    answer(exclude(states(all), border(texas)))
\end{align*}
Dependency based Compositional Semantics(DCS) \citep{liang2013-Just-DCS} is another tree-structured logical forms, where the logical-form, a tree, is called DCS Trees. In its basic version DCS proposes only two operations \textit{join} and \textit{aggregate} and the full version comes with higher order function like \textit{argmax} etc. readers are referred to \citet{liang2013-Just-DCS}. An example of DCS tree as logical form is shown in figure \ref{fig:dcs_tree}.
DCS tree proposes to reduce the complexity in compositinally creating the logical form of a sentence, however, its being a tree-structured logical form brings some limitations, such as it can't be used to represent bound anaphora as in sentence ``those who had a child who influenced them" \citep{liang2013lambda-DCS}. 

\section{Logic Based Formalism}
\label{sec:logical-form}
Many semantic parsing systems use higher-order formal logic to represent meaning of natural language sentence e.g. $\lambda-$calculus \citep{zettlemoyer2005ccg}, $\lambda-$DCS \citep{berant2013semantic, berant2014pp}. A first order predicate logic can only express simple natural language sentence of type yes/no or one that seek set of elements fulfilling a logical expression. To operate on set of elements the formal languages are augmented with higer-order function, eg. count(A) that would return cardinality of set A.  $\lambda-$Calculus \citet{carpenter1997LambdaCalculus} is a higher order functional language, it is more expressive, it can represent natural language constructs like count, superlative etc. $\lambda-$DCS \citep{liang2013lambda-DCS} which doesn't use existential variable, borrows hugely from $\lambda-$Calculus. Logical formalism that was introduced by \citet{liang2013-Just-DCS} as DCS-tree, will be discussed in Section \ref{sec:tree-form}

\subsection{Statistical Models}
\citet{zettlemoyer2005ccg} have used $\lambda-$Calculus as intermediate logical form to represent meaning of the natural language sentence. The construction mechanism requires an initial set of CCG lexicon $\Lambda_0$, where the lexicons have also been assigned semantic meaning using $\lambda-$abstractions, e.g. $borders := (S\backslash NP)/NP : \lambda x.\lambda y.borders(y, x)$  for sentence ``Utah borders Idaho". Given a sentence and its logical-expression, a rule based function $GENLEX(S_{i}, L_{i})$ creates  lexicon specific to the sentence. Together with $\Lambda_0$, the new set $\lambda = \Lambda_{0}\cup GENLEX(S_{i}, L_{i})$  form the search space for the probabilistic CCG parser. The parser uses beam-search to come up with a high probability parse of the sentence. There are sequence of derivation stages $T_i$ used by a parser to reach to the logical form $L_i$, however they are taken as hidden variable while learning the parameter of the parser. The set of lexicon $\Lambda$ will swell when trained over larger datasets, even though the algorithm adds only lexicon that are part of the final logical expression.  This work was state of the art at the time, giving precision of $96.25\%$ and recall $79.29\%$ on Geo880 dataset.  

\citet{berant2013semantic} was perhaps the first work using a simplified $\lambda-$DCS as the intermediate logical form, there system is called SEMPRE. The logical form is built recursively in a bottom up fashion starting with a lexicon leading upto the final logical form. The lexicon are logical predicates in KG which are used for a natural language phrase. The process of determining a lexicon involves creating typed-NL phrases $R_{1}[t_{1}, t_{2}]$ by aligning a text-corpus with KG \citep{lin2012TypedPhrases}, then the set of supporting entity-pairs $\mathcal{F}(R_1)$ is determined, the logical predicate $R_2$ makes for a lexicon if $\mathcal{F}(R_1)\cap\mathcal{F}(R_2)\neq\phi$. Using typed-phrases also helps in tackling polysemy e.g. \textit{"born in"} could mean \texttt{PlaceOfBirth} or \texttt{DateOfBirth} which can be tackled using typed \textit{"born in"}\texttt{[Person, Location]} and \textit{"born in"}\texttt{[Person, Date]} respectively. For composition of lexicon into logical form there are small set of rules: say $z_{1}, z_{2}$ are two lexicons, they can undergo following operations join $z_{1}.z_{2}$, aggregations $z_{1}(z_{2})$, intersections $z_{1}\cap z_{2}$ or bridging $z_{1}\cap p.z_{2}$. The last operation bridging is used when the relation between two entity is weakly implied or it is implicit. If the two entities(also called unary $z$) have types, say $t_{1}, t_{2}$ bridging introduces all binary predicates $p$ that have type structure $[t_{1}, t_{2}]$, thus the logical form $z_{1}\cap p.z_{2}$. During the process of generating the logical form, a feature vector take shape as well, which is later used to score the candidate logical form with a log-linear model. During evaluation the logical-form which is most probable $p_{\theta}(d|x)$ gets used. SEMPRE scores F1 $32.9\%$ on WebQuestions which \citet{berant2013semantic} introduced. 

\paragraph{Discussion} Of the two systems described above, \cite{zettlemoyer2005ccg} have used $\lambda-$Calculus with additional quantifiers count, argmax, and definite operator besides the universal and existential quantifiers. However, \citet{berant2013semantic} have tried to simplify the verbosity of $\lambda-$expression by not using existential quantifiers, considering them implicit, and borrow other operators of $\lambda-$Calculus when necessary. Qualitatively both are logical forms are equally expressive. The difference between the two approaches are in the way they carry out the learning of the parser. \citeauthor{zettlemoyer2005ccg} uses a fully supervised approach, by using a fully annotated dataset of sentence and logical expression, while \citeauthor{berant2013semantic} uses weak-supervision by using a dataset of question and answer pairs. We can't compare these to model empirically here because they have been applied on two different dataset. 

\subsection{Neural Encoder-Decoder Models}
A neural encoder-decoder architecture can  learn to generate formal logical expression representing meaning of a sentence. There are many works employing ecoder-decoder architecture in the context of semantic parsing e.g. \cite{dong2016Seq2TreeAndSeq, jia2016seq2seq_data_augmentation, dong2018coarse2fine}. Unlike statistical models, described above which require feature engineering and high quality lexicon \citep{zettlemoyer2005ccg}, the Neural Network based model can be trained to learn the features required for semantic parsing by themselves. 

\begin{figure}[h]
    \centering
    \includegraphics[scale=0.4]{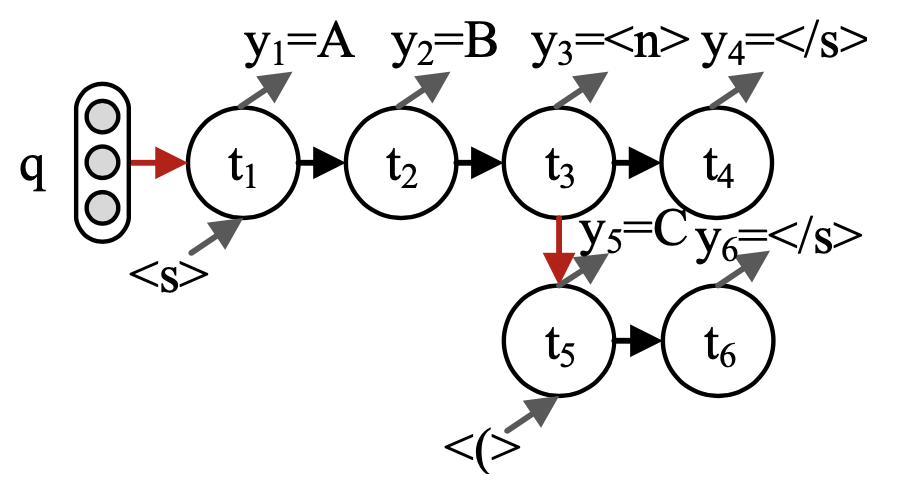}
    \caption{tree-structure for logical form AB(C)}
    \label{fig:tree_abc}
\end{figure}

\citet{dong2016Seq2TreeAndSeq} have considered the problem of semantic parsing as sequence transduction task, converting a sequence of words in natural language into a sequence of $\lambda$-terms in the logical form. The  encoder-decoder  architecture is made of an LSTM-encoder and an LSTM-decoder. The index of the words in NL sentence $q=x_{1}\dots x_{\abs{q}}$ is first converted to a vector using an embedding matrix, then passed to encoder's input layer $(l=0)$, one-by-one in sequence until time step $t_{\abs{q}}$, when encoder has seen all the words in the input sentence. Next time step and onward belongs to the decoder. The input to the decoder in its first time step is hidden state of encoder $h^{L}_{t_{\abs{q}}}$ and the index of the start of sequence symbol $\langle s \rangle$. At each time step afterward the decoder takes at input layer $(l_0) $ word vector corresponding to the previous predicted word $y_{t-1}$ and gives the probability $p(y_{t}|y_{<t}, q)$ distribution on output vocabulary of logical tokens/predicates. At inference the next word in the output sequence is obtained using a greedy-search (first-best) over output vocabulary and its conditional probability returned by the model $p(y_{t}|y_{<t}, q)$. The seq2tree architecture proposed by them considers the hierarchical nature of the logical form. It generate a tree-structured logical form recursively, figure \ref{fig:tree_abc}. They have evaluated their systems in closed domain on Geo880 dataset, where the seq2tree models scores F1 $87.1\%$

\citet{jia2016seq2seq_data_augmentation} as well have employed encoder-decoder architecture and train their model on augmented dataset. The augmented dataset is obtained using production rules of context free grammar, e.g. replacing an entity by its type, or replacing a word of, say type $t_1$ by a whole-phrase when the phrase type checks out $t_1$. 

\citet{dong2018coarse2fine} improves upon its previous work slightly by introducing another intermediate decoder called sketch-decoder. Thus it has for modules in the network: $\text{input-encoder}\rightarrow \text{sketch-decoder}\rightarrow \text{sketch-encoder}\rightarrow \text{output-decoder}$. The idea being to gloss over the low-level information like variable names, their values to create a coarse-representation of the input sentence, which can they guide the output-decoder into generating better formed logical expressions.

\paragraph{Discussion:} LSTMs are easy engineering tools for sequence-transduction, but they don't learn the rules of the grammar \citep{sennhauser2018lstm-fail_in_rules}. \citet{sennhauser2018lstm-fail_in_rules} used 4 production rules and generated a 2Dyck language, which is a string having only brackets `['  `]' and `\{'  `\}',  e.g. string ``\{[\{\}[]]\} ". They showed that LSTM don't generalise well, the error rate for out-of-sample test data is 8-14 times high compared to in-sample error-rate of 0.3\% for LSTM using 50 hidden units. They observe that LSTM learn the sequential-statistical correlation and that they don't store irrelevant information, but LSTM fail to learn the 4-rules of the CFG used to generate the language. They observe that novel architecture may be required to learn the grammar-rules, and there are works in that directions e.g. dynamic network architecture \citep{looks2017dynamic_computation_graph} of hardwired structural constrain \citep{Kiperwasser_2016_hardwaired_structure, arm2015inferring_hardwired_structure}

\section{Graph based formalism} 
\label{sec:graph-form}
The semantic parser using graph based formalism resort to a labeled-graph also called Semantic Query Graph(SQG) as semantic representation of the natural language (NL). Graph as an abstraction is widely used and Label graph are a good way to represent the meaning of NL sentence, where the nodes represent a concept/person and the edges between represent their inter-dependency. Besides, graph is also the language of many Knowledge Graphs(KG). There are many works using SQG as logical formalism liek \citet{reddy-etal-2014-large, Scott_yih_and_Jianfeng_2015_SQG-Tx-KB, bao2016ComplexQuestions, sen_hu_ieee-2018_NFF, sen_hu_emnlp_2018SQG-Tx-KB}.

\paragraph{Semantic Query Graph:} A semantic query graph SQG, by \citet{sen_hu_ieee-2018_NFF}, is a graph, in which each vertex $v_i$ is associated with an entity phrase, class phrase or wild-cards in the natural language (NL) sentence ; and each edge $e_{ij}$ is associated with a relation phrase in the NL sentence, where $1 \leq i, j \leq |V|$.

The semantic parsing pipeline used by \citet{reddy-etal-2014-large} takes a natural language (NL) sentence and uses a syntactic parser for CCG by \citet{clark-curran2004-ccg-parser}. A CCG grammar provides with one-to-one correspondence between syntax and semantic, syntax being the word-category and semantic represented using a $\lambda$-expression, together they are referred as CCG-lexicon. Further in the pipeline there is a graph-parser parsing the $\lambda$-expression as a label graph. The nodes in the graph may be attached to math functions like unique, count etc. as required by the semantics of the NL sentence. The label-graph is not yet contextualized against a world/KG, therefore also called an ungrounded graph. A contextualization process of nodes and edges of the ungrounded graphs against a KG is carried out using beam search. So, given an ungrounded graph (u), labels on its nodes and edges are run through KG to find possible candidate entities and relations respectively. Thus obtaining a set of possible grounded graphs \{g\}. A structured-perceptron is then used to wade through candidate graphs by scoring their feature vectors $\bm{\Phi}$ as dot product with model parameters $\bm{\theta}$.

\citet{Scott_yih_and_Jianfeng_2015_SQG-Tx-KB} have tried to bring in a part of $\lambda$-calculus into a graphical representation by introducing variable nodes (corresponding to existential variables) and answer nodes (corresponding to the bounded variable in lambda-expression), as well as aggregation function such as argmin, argmax, count into semantic query graph. The generation of a query graph is formulated as transition of states, each state being a subgraph in the KG thus always grounded. Starting with an empty query graph $\phi$ to state $S_{e}$ with only entity nodes, the next state add path to answer node known as state $S_p$ finally adding the constraints to obtain the full query graph as state $S_c$. The transition from one state to another is defined in terms of a well defined set of actions $\mathcal{A}=\cup \{\mathcal{A}_e, \mathcal{A}_p, \mathcal{A}_c, \mathcal{A}_a\}$ i.e. add entity nodes, path nodes, constraints nodes and aggregation respectively. Candidate $S_e$ states, which only contains single entity nodes, are decided using an entity linking system \citep{yang2016smart_entity_linking}. Transitioning to state $S_p$ requires making 1-hops/2-hops and scoring all the candidate paths for a possible predicate by comparing its similarity with questions-pattern (obtained from the question after replacing the entity in $S_e$ with place holders <e>) using two CNN models. The construction of the final SQG which is state $S_c$ requires adding constraints (entity/class) or math functions, which is guided by heuristics, e.g. of all the resource-nodes in KG attached to variable nodes in $S_p$ select one if it is an entity occurring in the question. The candidate SQG thus obtained are ranked by the F1 score based on answers they receive when executed on KG.

\citeauthor{sen_hu_ieee-2018_NFF} introduced two frameworks, RFF (Relation (edge)-First Framework) and NFF (Node First Framework), for semantic parsing capitalizing upon semantic query graph (SQG). The constructions mechanism in both the framework uses dependency parse of a sentence, say $Y(N)$. In RFF the relation mentions is a subtree $y$ in $Y(N)$. The relation mention should have a matching relation in a predefined  set $T=\{rel_{1}, rel_{2}, \dots, rel_{n} \}$. A matching relation $rel_i$ must have all its words in the subtree $y$. The associated nodes/arguments to relation edge $\langle rel_{i}, arg1, arg2\rangle$  are determined based on pos-tags coming out from the subtree $y$. Further, the relation mentions and the node/arg-phrases, which are in their surface form, are mapped to predicates/predicate paths and entities/classes respectively in the KG. The mapping from $rel_i$ to predicates/predicate paths uses a relation mention dictionary $D^R$, which utilizes the set of relations mentions and their Supporting-Entities as in Patty \citep{nakashole2012patty}. Similarly, the mapping from nodes/args-phrase uses an entity mentions dictionary $(D^E)$ CrossWikis by \citet{spitkovsky2012CrossWikis}.
The second framework NFF is pitted to be robust to errors in dependency parsing as compared to RFF. RFF uses dependency-tree to determine relation mentions, dependency structure and pos-tags, however NFF uses dependency-tree only to decide if there should be an edge between to entity nodes. Another advantage of using NFF is that it allows unlabelled edges between entities to represent implicit relations eg. an unlabelled edge between nodes Chinese and actor. The implicit relations are resolved during query evaluations. NFF requires first to extract all entity mention using a dictionary-based approach \citet{deng2015Entity-Dict}, then using dependency tree introduce edges between nodes: wherever two entity-mentions are adjacent in the dependency-tree the relation edges in the SQG is kept unlabelled (implicit relation) and where they are far apart, the words in the dependency-tree makes for the label of the relations edge in the SQG.
The query evaluation in both the framework is similar i.e. finding top-k matching subgraph in KG corresponding to a SQG. Each node and edge in the SQG comes with a candidate-list and each candidate gets a tf-idf score when retrieved from $D^{R}/D^{E}$. The only difference being SQG in the NFF may have few edges left unmatched. \citet{sen_hu_ieee-2018_NFF} also propose to use bottom-up approach of forming the SQG and finding a correct match. Starting with a single node of the SQG and finding a match in the KG, then expanding the node to a partial SQG and scoring its corresponding match. 

\citet{sun2020sparqa} allude to the fact that current dependency parser err in longer and complex sentences and they propose to first do a coarse (skeleton) parsing of the complex sentence into auxiliary clauses and pass it over to NFF \citep{sen_hu_ieee-2018_NFF} to take up the fine-grained semantic parsing. The skeleton parsing is modeled as set of four steps 1. identifying whether a sentence could be parsed into main-clause and auxiliary-clause, 2. identifying the text-span in the sentence that makes for an auxiliary clause after the first step is true, 3. identifying the headword in the sentence that governs the texts-span and 4. identifying the dependency relation between headword and the text-span. The four above steps are modelled using BERT \citep{devlin2019bert}. Four different $BERT_{BASE}$ models are fine-tuned as a task in single sentence classification (SSC), question answering(QA), questions answering and sentence pair classification (SPC) respectively for step 1, 2, 3 and 4. Different from NFF which uses tf-idf score for ranking candidates and the matched query graph, \citeauthor{sun2020sparqa} have used a sentence-level scorer and a word-level scorer. The sentence-level scorer scores the similarity of a test-sentence against a training-sentence, favoring training sentences which can provide their underlying query graph to the test-sentence which when executed on the KG should retrieve non-empty result. The sentence-level scorer is a fine-tuned BERT for the task of SPC. The word-level scorers scores bag-of-words (BOW) in sentence (after removing entities and stop words) and BOW in query graph which mainly consists of predicates, thus scoring appropriateness of the predicates used in the SQG. 

\paragraph{Discussion:}
We summarized four semantic parsing systems above, which use graph as intermediate logical form. They differ mainly in the way they generate the SQG, while \citet{reddy-etal-2014-large} and \citet{sen_hu_ieee-2018_NFF} use a syntactic parser (CCG and dependency-tree) \citet{Scott_yih_and_Jianfeng_2015_SQG-Tx-KB} doesn't use any suntactic parser. It uses a state-transition based approach to generate the SQG keeping partial states always grounded. We see that on WebQuestions \citep{Scott_yih_and_Jianfeng_2015_SQG-Tx-KB} STAGG achieves $52.5\% F1$. The state-transition based approach with a different set of actions (connect, merge, expand and fold) is used in a different work by Sen Hu \cite{sen_hu_emnlp_2018SQG-Tx-KB} which got an F1 of $53.6\%$ on WebQuestions.

\section{Tree structured logical form}
\label{sec:tree-form}
\citet{cheng-reddy-2017-tree_transformation_rule} uses nested tree-structured logical form known as funQL. FunQL has a predicate-argument structure, where the predicate is a non-terminal (NT) and the arguments may be another sub-tree or a terminal node. The s-expression in funQL besides telling how the semantics are composed also tells how it could be derived by nesting words of natural language taken either as terminal or non-terminal node of the tree. The semantic parsing system by \citet{cheng-reddy-2017-tree_transformation_rule} separate the generation of logical form and mapping of lexicon to knowledge base entity and relation into two different stages. The logical form generation is done in a task-independent fashion. The logical form obtained is called ungrounded logical form. The model generates the ungrounded logical form by applying sequence of actions in actions-set $A=\{NT, TER, RED\}$ on the stack of s-expression, and choosing a word from an input-buffer (used to store words in the sentence) randomly. The model is trained to learn a distribution of possible action set and the word to be chosen from the input buffer conditioned on the sentence and the state of the stack of s-expression. The stack is represented using a stack-LSTM \citep{dyer2015stack-LSTM}, say $(s_t)$. The state of input-buffer is encoded using a Bi-LSTM \citep{hochreiter1997LSTM} and is adaptively weighted at each time step using the stack-state $s_t$. The two probabilities are given below:
\begin{align*}
    p(a_t |a_{<t}, x) \propto \exp(W_a\cdot e_t)\\
    p(u_t|a_{<t} ,x) \propto \exp(W_s \cdot s_t) \\
\end{align*}
where $e_t$ is concatenation of $b_t$ and  $s_t$ and $W_a, W_s$ are weight matrix. The mapping of lexicon to database entity and predicates is done using a bi-linear neural network 
\begin{align*}
p(g_t|u_t) \propto \exp \bm{u_t} \cdot W_{ug} \cdot \bm{g_t}^T
\end{align*}
The training objective is to maximize the likelihood of the grounded logical form, therefore it considers the ungrounded logical form as latent variable and marginalise over it.

\paragraph{Discussion} The transition based approach with just tree-operation can generate all the tree-structured candidate logical-form. It makes a good case for its adoptions where the inherent structure of the language could be limited to tree-like.  

\section{Benchmark Datasets}
\label{sec:benchmark}
The benchmark dataset used to evaluate the few systems discussed in this survey are described in table \ref{benchmark_dataset}. Benchmark datasets have grown in complexity as well as in size overtime. Partly driven by demand of larger and larger machine learning model which require large data to train and partly to find a better semantic parser which would be able to parse complex and varied set of questions. 

\paragraph{Geo880:} Introduced by \citep{zelle1996learning} is a closed domain dataset of questions related to US geography. Semantic parsing systems based on encoder-decoder architecture seems to favour close domain dataset. Thus purely evaluating the ability of the semantic parser to generate correct logical form.

\paragraph{WebQuestions} \citep{berant2013semantic} introduced this dataset, collected from google suggest api. The dataset has question and denotation paris, collected from Freebase\citep{bollacker2008freebase}.  \cite{yih2016_WebQSP_ds} later annotated the data with logical forms, to show that annotation help in learning.   

\paragraph{GraphQuestions} \citep{su2016GraphQuestion} introduced this dataset which was obtained by presenting Aamazon Mechanical Turk workers with 500 Freebase graph queries and asking them to verbalise it into natural language. 

\paragraph{ComplexWebQuestions} ComplexWebQuestions v1.1 \citep{talmor2018CWQ_v1p1_ds} is a dataset of $34689$ complex questions split into train test and dev set. Each question comes with a SPARQL query that can be executed against Freebase \citep{bollacker2008freebase} as well as a set of web-snippet (366.8 snippet per question) which can be used by a reading comprehension model to find out an answer of the questions. The dataset was created using seed from WebQuestionsSP \citep{yih2016_WebQSP_ds}, where a seed SPARQL query is taken from the WebQuestionsSP and combined with another fact from freebase according to some set rules. The SPARQL query thus formed is complex, which is then automatically translated in a natural language using template. The question so formed is not correct but could be understood, to get a grammatically correct form of the question Amazon Mechanical Turk workers are asked to paraphrase it.

\paragraph{ComQA} The questions in ComQA \citep{abujabal2018comqa_ds} come from WikiAnswers Community QA Platform. The 11,214 question are divided into 4,834 clusters of paraphrases with help from crowd-sourcing. The questions are real and not based on templates. There are variety of questions such as simple, temporal, compositional (requiring answer of simple parts first before the final answer to the questions is possible), comparison (comparative, superlative and ordinal), telegraphic (keyword-queries), tuple (connected entities form an answer) and empty(questions with no answers).

\begin{table*}[htb]
\centering
\begin{tabularx}{\linewidth}{l*{3}{R}}
    \hline
    \leftHead{Dataset} & \leftHead{Source} & \leftHead{Pairs} & \leftHead{Train-Test-Dev} \\\hline
    Geo880 \citep{zelle1996learning}&  -- & Q-LF  & 600, -, 280 \\
    WebQuestions \citep{berant2013semantic}& GoogleSuggest  & Q-Ans  &  5810, -, - \\
    GraphQuestions \citep{su2016GraphQuestion}& FB & Q-LF & 5166, - , -\\
    ComplexWebQuestions \citep{talmor2018CWQ_v1p1_ds}& WikiAnswers &  & 34689,-,-\\
    ComQA \citep{abujabal2018comqa_ds}& WikiAnswers &  Q-LF& 11214, -, - \\
    \hline
\end{tabularx}
    \caption{Benchmark Dataset}
    \label{benchmark_dataset}
\end{table*}


\bibliography{acl2020}
\bibliographystyle{acl_natbib}



\end{document}